# Technical Challenges for Ultra Fine Pitch Hybrid Bonding

Charles G. Woychik†, Jonathan Abdilla††, Chris Scanlan††, Victor Vilar†, Robert Patti†, Dadi Setiadi†, and Himanandhan Reddy Kottur*, Navid Asadizanjani*

†NHanced Semiconductors, Inc.;
cwoychik@nhanced-semi.com

††Besi Austria GmbH;
*Department of Electrical and Computer Engineering, University of Florida

## Abstract

Hybrid bonding is rapidly emerging as a foundational technology for next-generation advanced packaging, enabling ultra-high-density interconnects required for heterogeneous integration of chiplets, logic, memory, and specialized accelerators. As the industry pushes toward ultra-fine pitch interconnects approaching single-digit microns and below, significant technical challenges must be addressed across the process flow. These challenges include achieving extremely tight wafer and die alignment tolerances, maintaining surface planarity and cleanliness at the nanometer scale, controlling copper dishing and dielectric erosion during CMP, and ensuring defect-free oxide and copper-copper bonding interfaces. In addition, scaling hybrid bonding to ultra-fine pitch requires improvements in metrology, defect inspection, thermal management, and yield control to support high-volume manufacturing. Integration with complementary technologies such as silicon interposers, wafer-to-wafer and die-to-wafer bonding architectures, and heterogeneous material stacks further increases process complexity. This presentation reviews the key technical barriers to ultra-fine pitch hybrid bonding and discusses emerging process solutions and equipment innovations required to enable reliable, high-density interconnects for future high-performance computing, AI, and advanced system-in-package applications.

## I. Introduction

Ultra-fine pitch hybrid bonding is emerging as one of the most important technologies enabling the next generation of semiconductor devices. As demand for higher performance, lower power consumption, and increased functionality continues to accelerate, traditional packaging and interconnect approaches are reaching their physical and electrical limits. Applications such as artificial intelligence, high-performance computing, advanced memory architectures, photonics, and heterogeneous integration require significantly higher interconnect density and shorter signal paths than conventional solder-based technologies can provide. Hybrid bonding addresses these limitations by combining direct copper-to-copper and dielectric-to-dielectric bonding at extremely fine pitches, enabling near-monolithic integration between chips, chiplets, wafers, and advanced substrates.

Despite its significant advantages, scaling hybrid bonding to ultra-fine pitches introduces a wide range of technical challenges across materials, process integration, equipment capability, and manufacturing reliability. As interconnect dimensions shrink to the micron and sub-micron regime, even minor variations in surface topography, particle contamination, alignment accuracy, or wafer warpage can lead to bond failures, electrical opens, or reduced yield. Achieving high-throughput manufacturing requires extremely tight control of chemical mechanical planarization (CMP), surface preparation, oxide activation, thermal expansion mismatch, and overlay precision. In addition, reliability concerns such as electromigration, thermal cycling fatigue, void formation, and stress-induced defects become increasingly critical as bond pitches continue to decrease.

The transition toward heterogeneous integration further increases the complexity of ultra-fine pitch hybrid bonding. Advanced systems now integrate logic, memory, photonics, RF devices, and specialized accelerators fabricated using different process technologies and materials. This creates new challenges in die-to-wafer (D2W) and wafer-to-wafer (W2W) bonding, thermal management, known-good-die strategies, testability, and supply chain scalability. At the same time, the semiconductor industry must balance performance improvements with manufacturability, cost, and long-term reliability to support high-volume production.

This paper examines the key technical barriers associated with ultra-fine pitch hybrid bonding and discusses the process innovations required to enable future advanced packaging architectures. Focus is placed on alignment accuracy, surface preparation, defect control, reliability, thermal considerations, and integration strategies that are critical for supporting next-generation heterogeneous integration platforms

## II. Changing Foundry Models

The Foundry 1.0 model refers to the traditional semiconductor manufacturing ecosystem in which integrated device manufacturers (IDMs) and dedicated semiconductor foundries operated with clearly separated roles. In this model, fabless semiconductor companies focused on chip design and outsourced wafer fabrication to pure-play foundries, enabling reduced capital investment and faster product development cycles. Foundry 1.0 emphasized high-volume silicon scaling driven by Moore's Law, with success largely dependent on lithography advances, transistor density improvements, and process node leadership. Packaging and assembly were typically treated as back-end commodity services performed by outsourced semiconductor assembly and test (OSAT) providers with limited system-level integration. While this model enabled rapid growth of the semiconductor industry for decades, it was primarily optimized for monolithic system-on-chip (SoC) designs and did not fully address emerging requirements for heterogeneous integration, chiplet architectures, advanced packaging, photonics integration, or system-level co-optimization that are now driving the transition toward more collaborative Foundry 2.0 approaches.

Foundry 2.0 is a manufacturing model that takes building blocks from traditional semiconductor manufacturers and uses advanced packaging and additive manufacturing to create highly customized devices. The Foundry 2.0 model is better suited for the future because it expands beyond traditional wafer fabrication to include advanced packaging, heterogeneous integration, chiplet assembly, photonics integration, and system-level co-optimization within a unified manufacturing ecosystem. As transistor scaling slows and the cost of leading-edge nodes continues to rise, future semiconductor performance gains will come from integrating multiple optimized dies through advanced packaging technologies such as hybrid bonding, silicon interposers, and high-density 3D integration. Foundry 2.0 enables closer collaboration among foundries, OSATs, substrate suppliers, EDA providers, and system designers to deliver complete platform-level solutions rather than isolated chips. This approach improves performance, power efficiency, design flexibility, and time-to-market while reducing development risk and enabling "best-of-class" integration of logic, memory, analog, RF, and photonics devices. The model is especially important for artificial intelligence, high-performance computing, defense, automotive, and data center applications, where system-level integration and advanced packaging have become as critical as transistor scaling itself.

Figure 1 illustrates the main reasons for this change in direction. Again, this relates to Gordon Moore's future direction, in which he states that "It may prove to be more economical to build large systems out of smaller functions which are separately packaged and interconnected" [1]. This is the future Moore's Law for the next generation of systems.

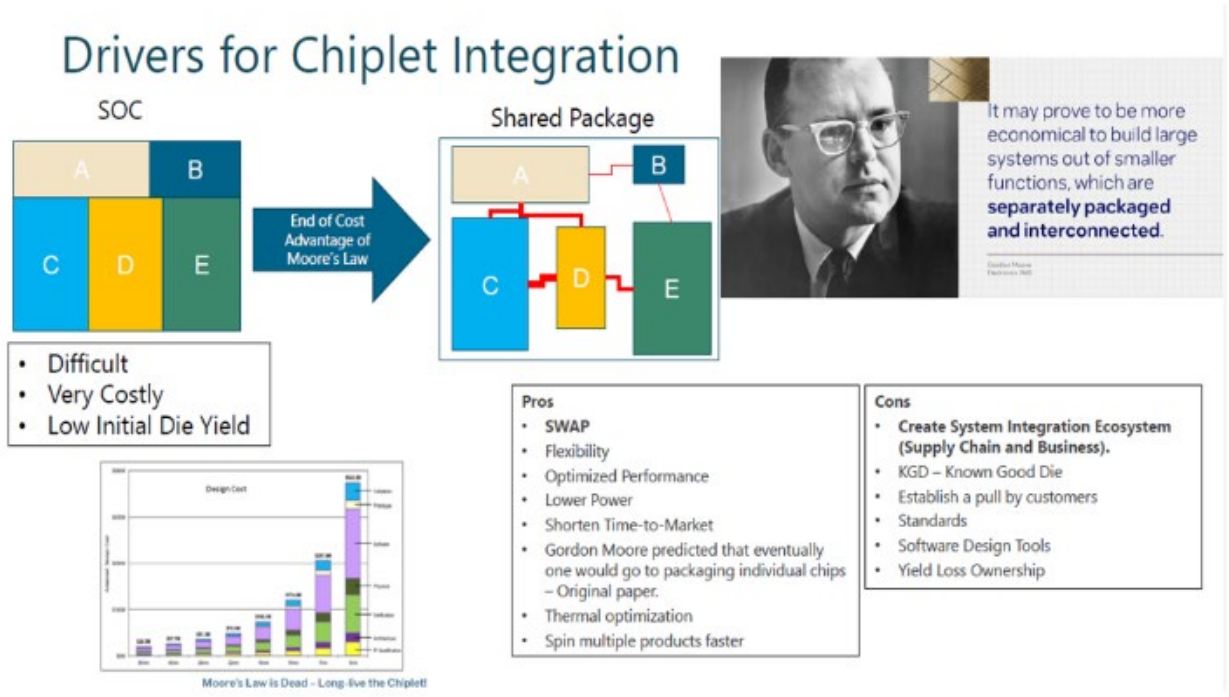

Figure 1: Technology drivers for disaggregation of a monolithic die into individual cells, called "Chiplets."

## III. Technical Drivers for Hybrid Bonding

As semiconductor devices continue to scale, the traditional approach of building a large monolithic system-on-chip (SoC) is becoming increasingly difficult and expensive due to lithography limits, lower manufacturing yield, rising power density, and escalating design complexity. To address these challenges, modern silicon devices are being disaggregated into smaller functional blocks known as chiplets. Each chiplet can be independently optimized for a specific function—such as compute, memory, analog, RF, or I/O—and fabricated using the most appropriate process node, improving both performance and cost efficiency. Chiplet-based architecture also improves manufacturing yield because smaller die sizes reduce the probability of defects impacting the entire system. In addition, advanced packaging technologies such as silicon interposers, 2.5D integration, 3D stacking, and hybrid bonding enable these chiplets to communicate with near-monolithic bandwidth and latency. This modular design approach accelerates product development, allows reuse of validated IP blocks, and provides greater flexibility for heterogeneous integration, making chiplets a critical enabling technology for next-generation high-performance computing, AI, photonics, and advanced semiconductor systems. Gordon Moore realized that this would eventually be the case [1]. Figure 1 illustrated how this approach yields a more optimized design at the lowest overall cost. It also emphasizes that semiconductor device node shrinking will continue, but only on the cells (chiplets) that require it. This is where packaging plays a critical role in providing the best Size-Weight-and-Performance (SWAP) solution. The problem currently is that the domestic capabilities need to be developed to accommodate this approach.

Figure 2 illustrates how device I/O pitch drives the appropriate interconnect technology [2]. The major takeaway is to use the appropriate technology for the contact density of the device. All these technologies will continue to be used; however, as the device pitch reaches 10 μm and less, hybrid bonding becomes the necessary technology that scales with decreasing I/O pitch.

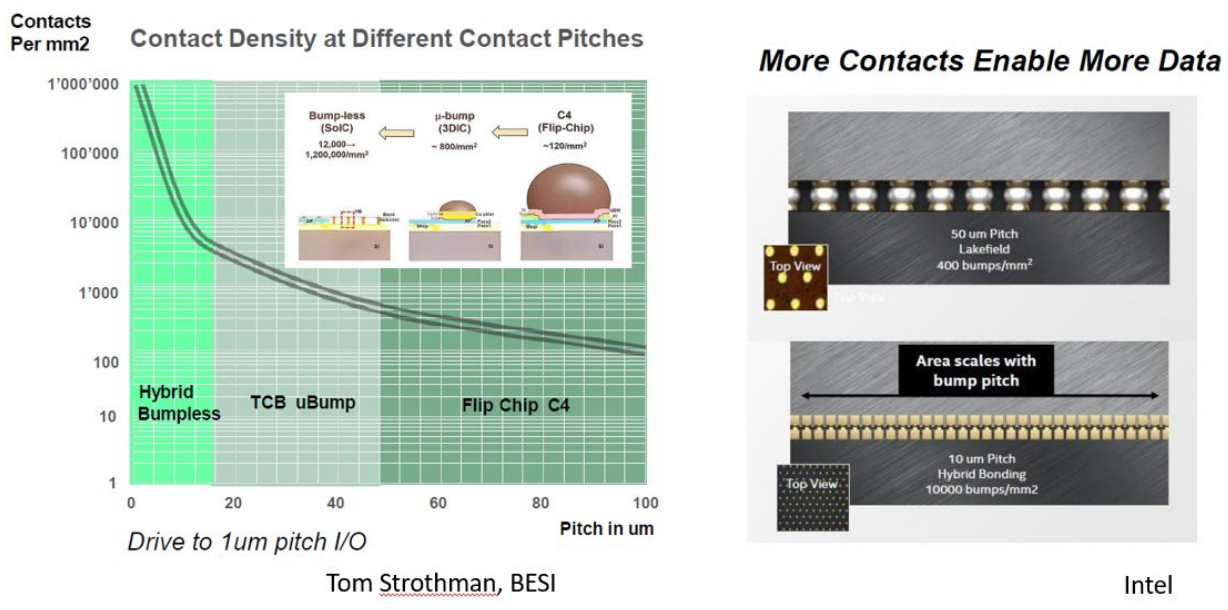

Figure 2: Comparison of the interconnect technology used for different ranges of I/O pitch.

The integration of chiplets requires a semiconductor back end of line (BEOL) capability. Figure 3 illustrates the evolution of packaging starting in the 1980s, when the semiconductor fab was completely independent of packaging. However, as time progressed, packaging required wafer bumping and test, which was still independent of the fab. As the device pitches decreased to sub 10 μm, it became necessary to implement hybrid bonding, which requires BEOL capability, and total design up front with the fab. This is where the future of packaging lies.

## IV. Mechanism for Hybrid Bonding

Hybrid bonding is an advanced semiconductor packaging technique that directly connects two wafers or dies using both dielectric-to-dielectric bonding and copper-to-copper interconnects. Unlike traditional solder-based micro-bumps, it enables extremely fine interconnect pitches, which significantly increases wiring density while reducing electrical resistance, power consumption, and signal delay. This makes it a key enabler for heterogeneous integration, 3D IC stacking, high-bandwidth memory (HBM), and advanced AI and photonics systems.

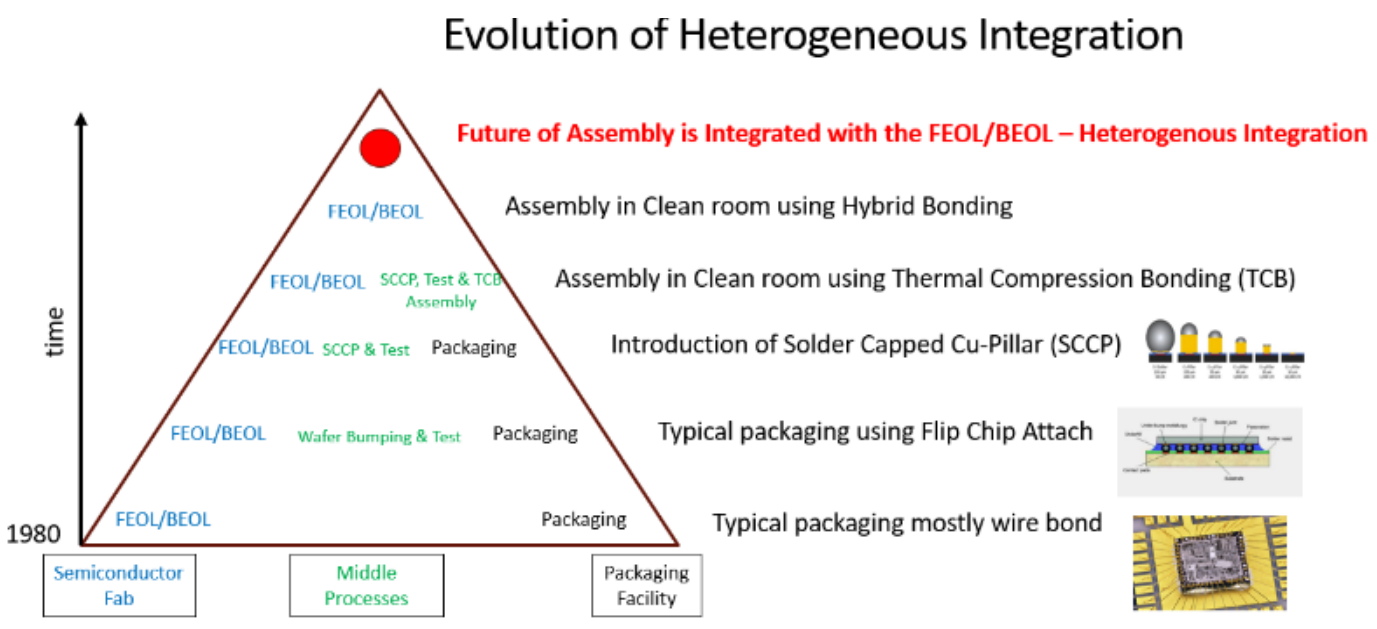

Figure 3: Evolution of packaging technology starting in 1980 to the present time.

The process begins by forming copper interconnects embedded in dielectric layers (typically silicon oxide) on each surface, followed by chemical mechanical polishing to achieve ultra-flat and smooth surfaces. After rigorous cleaning and surface activation (often plasma treatment), the wafers are precisely aligned at sub-micron accuracy. They are then brought into contact, where dielectric surfaces initially bond through weak intermolecular forces. A subsequent thermal anneal strengthens the interface and forms direct copper-to-copper electrical connections, completing both mechanical and electrical integration in a single step. Figure 4, provided by Adeia [3], schematically shows the three major steps that create a hybrid bond interconnect.

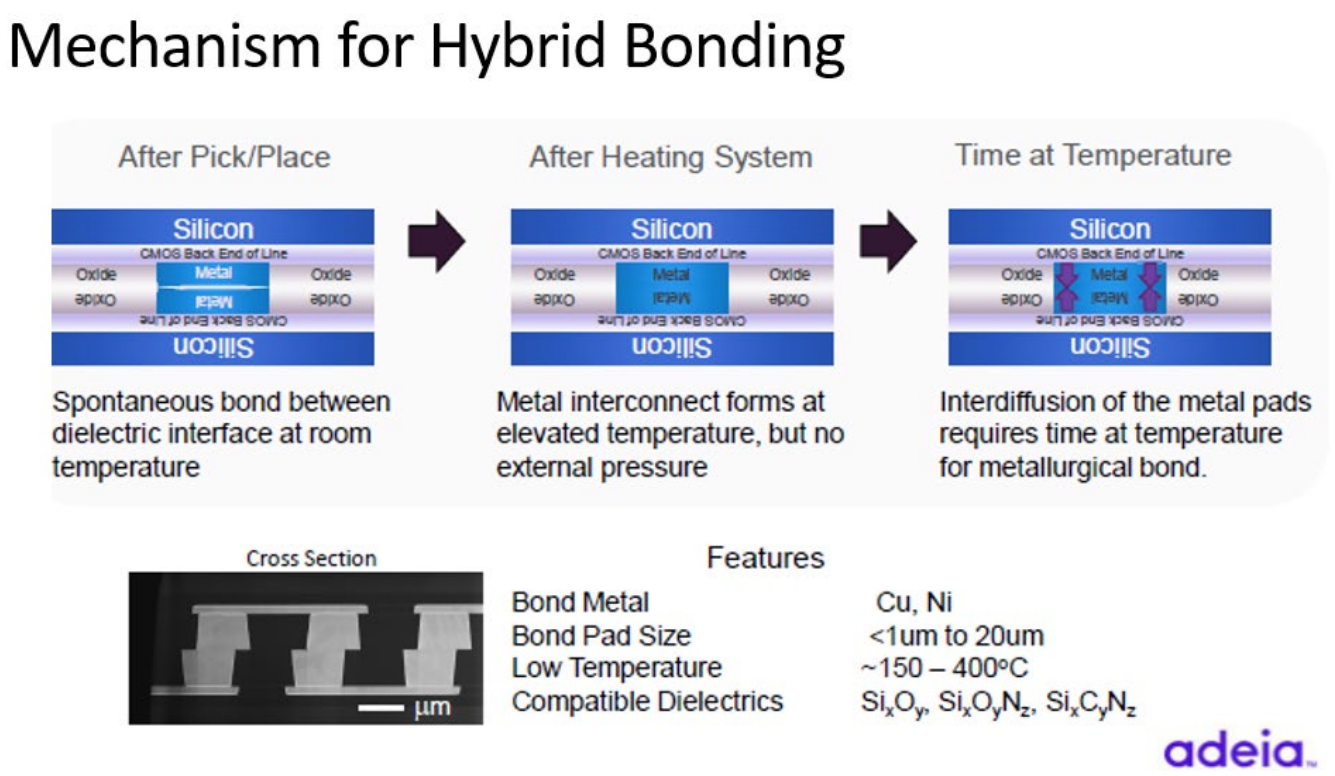

Figure 4: Main process steps used to create a hybrid bond [3].

Wafer-to-wafer (W2W) and die-to-wafer (D2W) hybrid bonding use the same fundamental mechanism—dielectric bonding plus direct copper-to-copper interconnects—but they differ in how the components are prepared and assembled, which has major implications for yield, flexibility, and cost.

In W2W hybrid bonding, two fully processed wafers are aligned and bonded across their entire surfaces in a single step. This approach offers the highest throughput and the best alignment accuracy because all features are processed in a uniform wafer-scale environment. It is especially attractive for applications where both wafers have similar die yields and identical layouts, such as certain types of memory stacking (e.g., HBM). However, its main limitation is yield-coupling: a defect on either wafer can potentially affect the entire bonded stack, which becomes more problematic as wafer sizes and defect densities increase.

D2W hybrid bonding, by contrast, involves picking and placing individual known-good dies (KGDs) onto a host wafer before bonding. This "chiplet-style" approach significantly improves overall yield because only tested and functional dies are integrated. It also provides much greater flexibility, enabling heterogeneous integration of different technologies—such as logic, memory, RF, or photonics—on a single platform. The tradeoff is increased process complexity, lower throughput, and more challenging alignment and placement control compared to W2W bonding.

In summary, W2W bonding prioritizes simplicity, alignment precision, and high-volume uniformity, while D2W bonding prioritizes optimization and system-level flexibility. As a result, W2W is often favored for homogeneous, memory-centric applications, while D2W is increasingly used for advanced chiplet-based architectures and heterogeneous integration in AI and high-performance computing systems. In this paper we will focus on the fundamentals of D2W bonding.

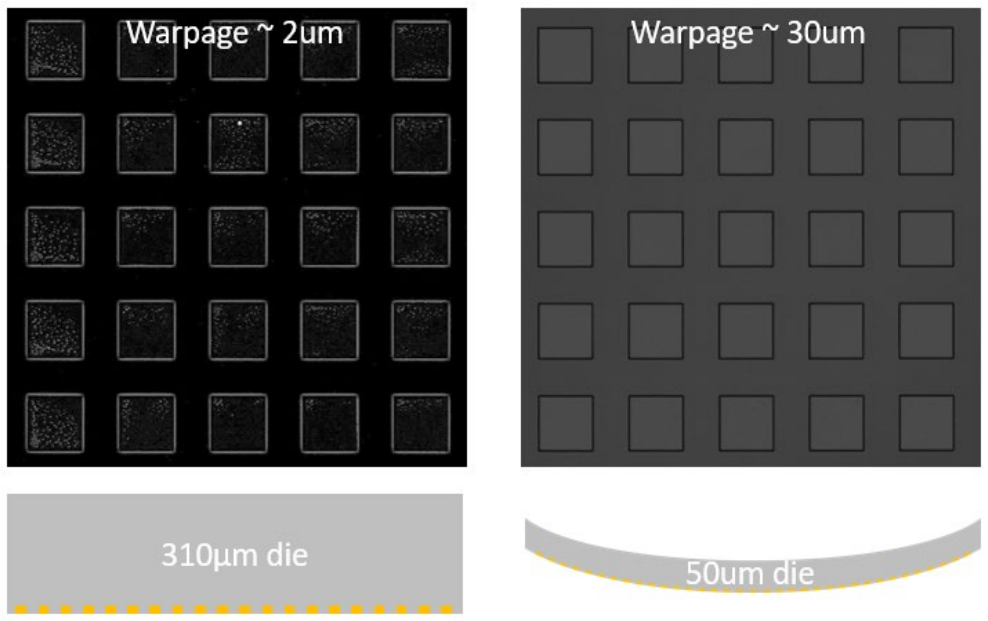

Figure 5: Impact of die thickness on shape forming.

Figure 5 illustrates the impact of die thickness on shape forming. Thicker dies have a tendency for lower intrinsic warpage, whereas thin dies can easily warp. The shape forming capability dramatically influences the quality of the bond. A thinner die (typically 50 µm or less) can easily be formed into a "smiley face" (convex) shape, where the initial contact with the mating surface is in the center and the weaker Van der Waals bonds migrate towards the perimeter of the die, producing a void free interface.

Die stacking enabled by hybrid bonding provides major gains in bandwidth and integration density, but it introduces significant manufacturing challenges, particularly related to micro-voiding and edge voiding at the bonding interface. Because hybrid bonding relies on direct dielectric-to-dielectric and copper-to-copper contact at extremely fine pitches, even nanometer-scale particles, surface roughness, or local non-uniformities can trap voids during initial contact and annealing. Micro-voids within the active bonding area can degrade electrical continuity, increase contact resistance, and create localized reliability risks under thermal cycling. Edge voiding is especially problematic because wafer bow, non-uniform CMP topography, and non-ideal pressure distribution during bonding often lead to incomplete contact initiation near the die or wafer periphery. These edge defects can propagate inward during annealing, reducing effective yield and limiting usable die area due to enlarged edge exclusion zones. As a result, stringent limits must be placed on surface preparation, planarization control, particle contamination, and alignment accuracy to ensure void-free bonding. While hybrid bonding enables unprecedented 3D integration performance, managing micro- and edge-void formation remains one of the key barriers to high-volume manufacturing scalability. Figure 6 shows the effect of die stacking on edge delamination.

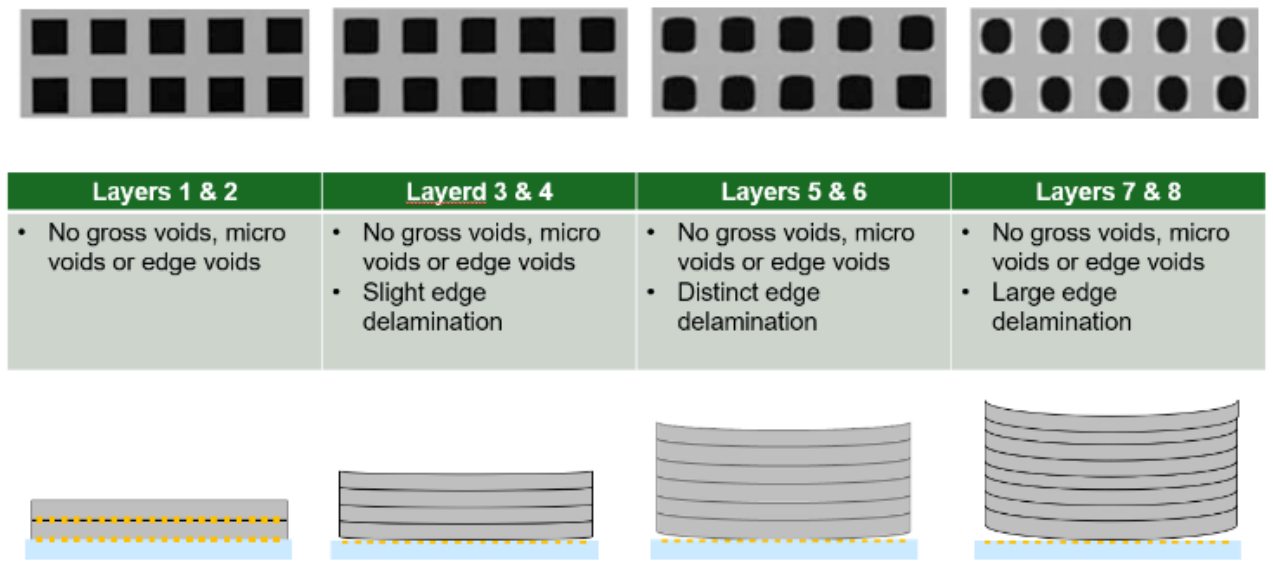

Figure 6: Die stacking challenges and how more layers affect edge delamination.

Die thickness and overlay (lateral alignment error between bonded dies) are two tightly coupled parameters that strongly influence yield and reliability in hybrid bonding. Die thickness variations affect global and local planarity during bonding; thinner dies are more flexible and can conform better to bonding surfaces, but they are also more susceptible to bow, warp, and handling-induced distortion. Thicker dies provide mechanical rigidity but can exacerbate non-uniform contact pressure, especially across large-area interfaces, increasing the risk of incomplete bonding initiation and localized micro- or edge-void formation. Overlay error, on the other hand, directly impacts the alignment of copper pads and dielectric openings, which are already at sub-micron or sub-100 nm pitch in advanced hybrid bonding schemes. Even small misalignment can reduce effective copper-to-copper contact area, increase electrical resistance, or lead to open circuits if pads partially miss their mating targets. When combined, thickness-induced warpage can effectively worsen apparent overlay across the die, meaning that local misalignment varies spatially even if global alignment is nominally within spec. This coupling makes simultaneous control of die thinning processes, stress management, and high-precision alignment systems critical. Ultimately, tighter control of both thickness uniformity and overlay is essential to minimize defectivity, improve bonding yield, and enable scaling to finer pitches and larger die sizes. Figure 7 shows the effect of die shape on assembly yield. Excessive tensile warpage resulting in a convex shape (smiley face – left side) results in the highest overlay error, as the die stretches. The other extreme is a compressive concave warpage (frowning face – right side) which yields the lowest overlay error, but results in void entrapment.

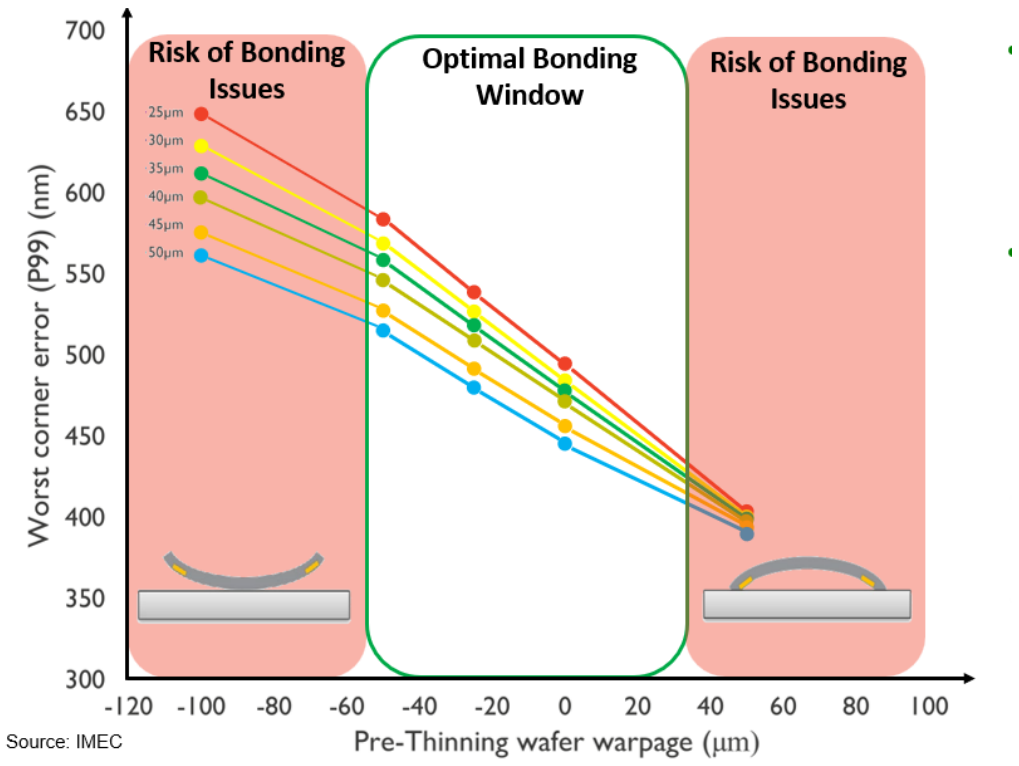

Figure 7: The effect of die shape on hybrid bond assembly yield.

Tool placement accuracy is one of the dominant factors determining yield in hybrid die bonding, particularly as interconnect pitches scale into the sub-micron and even nanometer regime. In hybrid bonding, bonding happens in sequence, mechanical first, electrical thereafter. Therefore, alignment accuracy (pick and place), and initial bonding energy (non-shifting) becomes challenging.

As placement accuracy improves, bond yield increases nonlinearly because misalignment directly impacts both electrical continuity and void-free interface formation. Even small lateral offsets can lead to incomplete copper overlap, increased contact resistance, or open circuits, while vertical and angular errors can exacerbate local pressure non-uniformity during bonding, increasing the likelihood of micro-voids and interfacial defects. Consequently, yield is highly sensitive to the combined error budget of the entire bonding system, including stage accuracy, thermal drift, die warpage, and pickup tool repeatability.

Modern hybrid-die bonders therefore rely on advanced optical alignment systems, in-situ metrology, and feedback-controlled stages to achieve sub-50 nm placement accuracy. However, as bump pitches continue to shrink, the allowable misalignment window narrows further, making yield increasingly constrained by tool capability rather than process chemistry. In practice, the relationship between placement accuracy and yield is steep: small improvements in alignment precision can produce disproportionately large gains in functional die-to-die connectivity and overall device yield, especially in high-density chiplet and 3D integration architectures. Figure 8 illustrates the key performance factors for die placement accuracy.

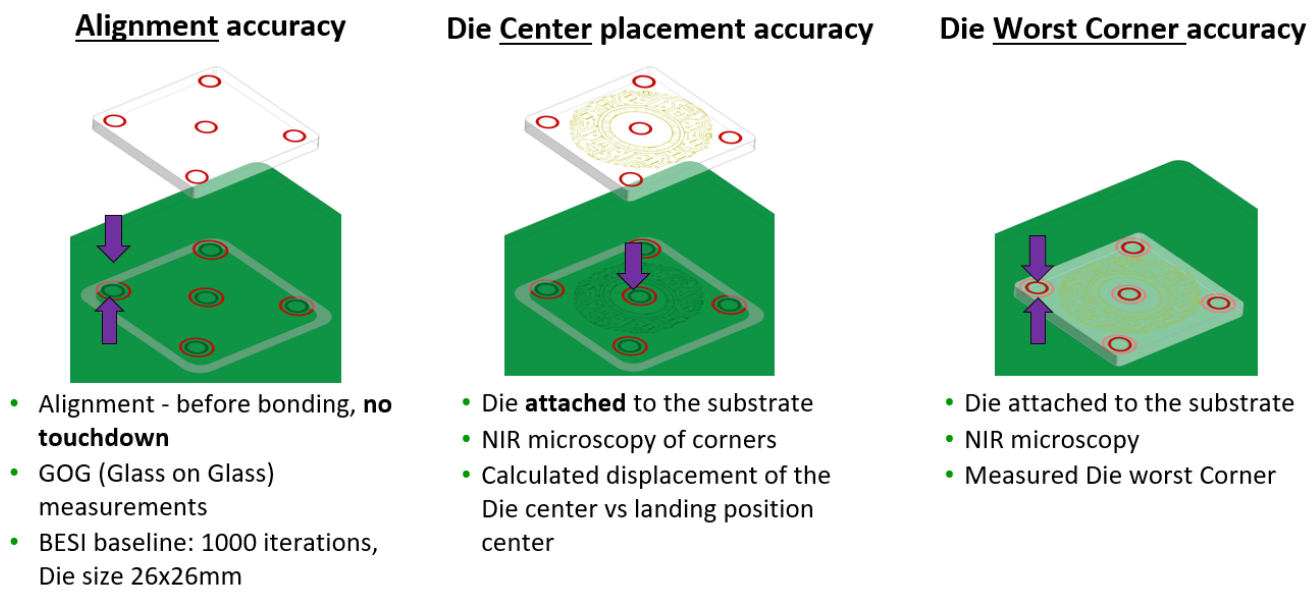

Figure 8: Key performance factors that influence die placement accuracy.

Some data have been taken on alignment accuracy while maintaining high throughput using the Besi Datacon 8800 Chameo bonder having the specifications of <200 nm overlay, for 100-300 mm host wafer support using a pick from film frame. The results for two different alignment methods are shown in Figure 9. The Circle approach shows a dramatic improvement of 55 nm in the x-direction and 76 nm in the y-direction. The Cross approach is still better than the specification, but not as good as the Circle approach.

Fine-pitch hybrid bond interconnects are a key enabler of next-generation 3D integration and chiplet architectures, allowing extremely high-density electrical and mechanical connections between stacked dies without traditional solder micro-bumps. These interconnects are typically defined by pitch scaling from the ~10 µm regime down to sub-micron dimensions, where direct copper-to-copper bonding and dielectric-to-dielectric fusion replace conventional bump-based assembly. Figure 10 shows several different hybrid bond configurations starting from 10 µm pitch and below.

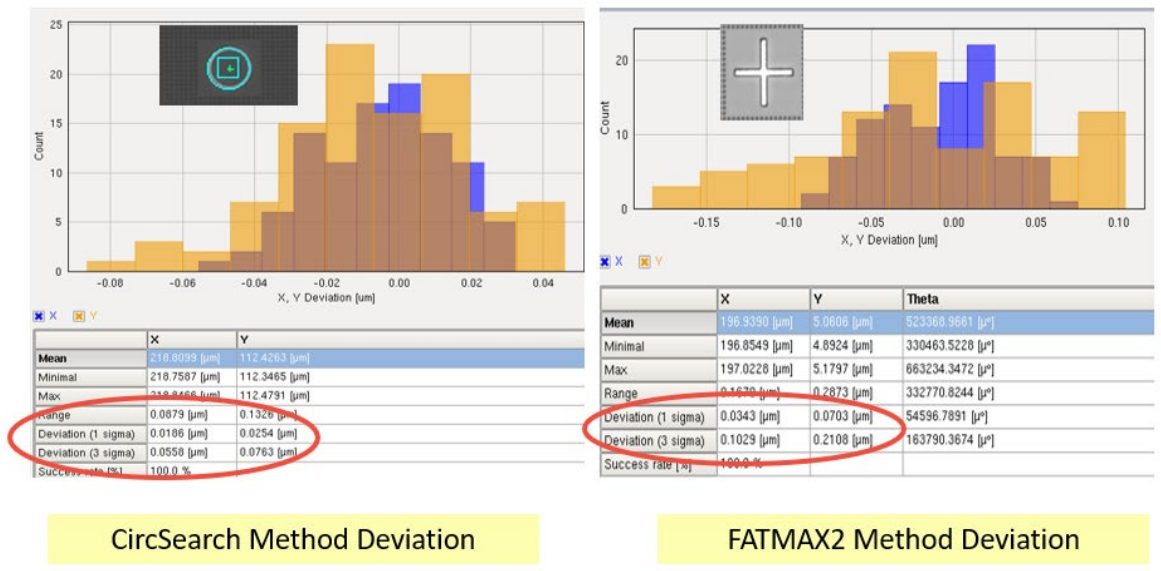

Figure 9: Comparison of two different alignment methods on impacting tool performance.

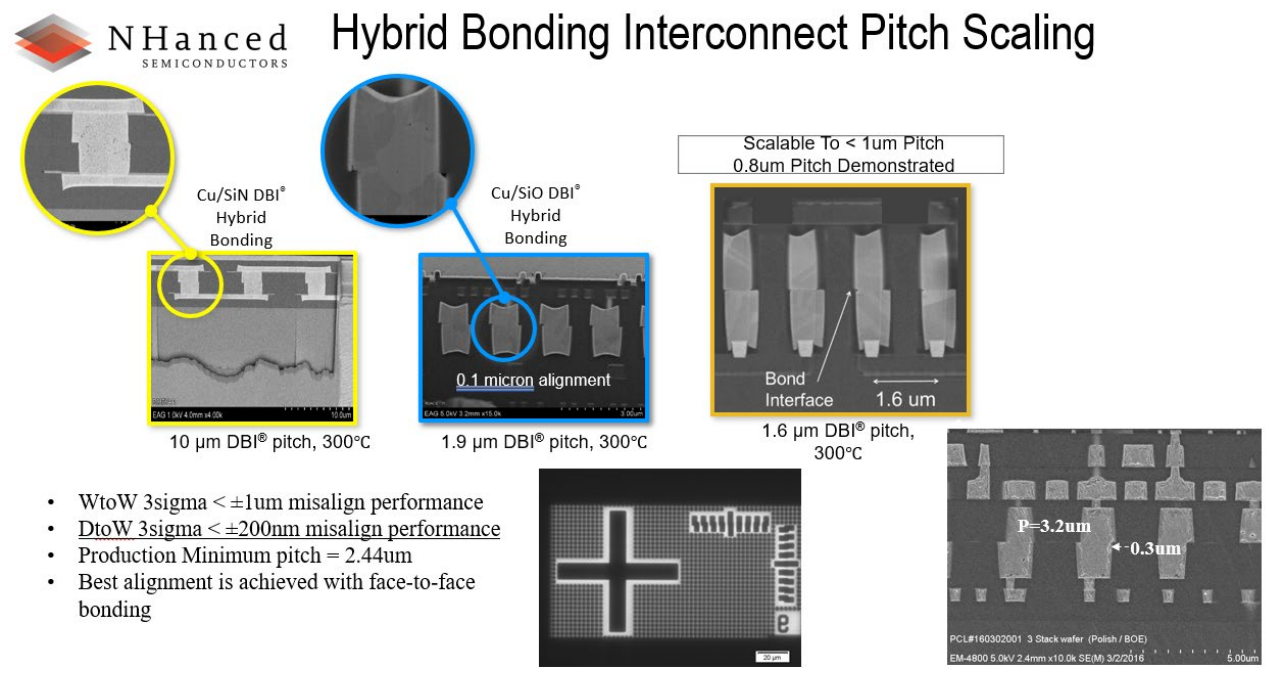

Figure 10: Examples of hybrid bond configurations ranging from 10 µm pitch down to 1.6 µm pitch (NHanced Image).

Beyond logic stacking, fine-pitch hybrid bonding is also emerging in memory and photonics integration. As pitch scales below ~5 µm and toward sub-micron experimental regimes, challenges such as wafer warpage control, oxide planarity, and copper surface preparation become dominant. At these dimensions, interconnect density approaches that of on-chip wiring, effectively blurring the boundary between intra-die and inter-die communication and making hybrid bonding a foundational technology for future heterogeneous integration systems.

Figure 11 shows a stack of eight 200 mm wafers, each wafer having a 14 µm thickness with 1.2 µm diameter TSVs. Complete bonding of all layers was achieved.

In addition, silicon and glass interposers having the cross-section shown in Figure 12 offer a packaging platform that can accommodate both solder and hybrid bond interconnections. For a Si-interposer, TSV dimensions can range from: 10 µm x 100 µm, 25 µm x 200 µm, and 50 µm x 300 µm. These same structures can be produced using a glass core. Both types of interposers can accommodate an 8-layer Cu Damascene structure, typically 4-layers on the top and 4-layers on the bottom. Typical metal dimensions are 2/2/2/2 µm L/W/S/T, but can down as low as 1/1/1/1 µm L/W/S/T.

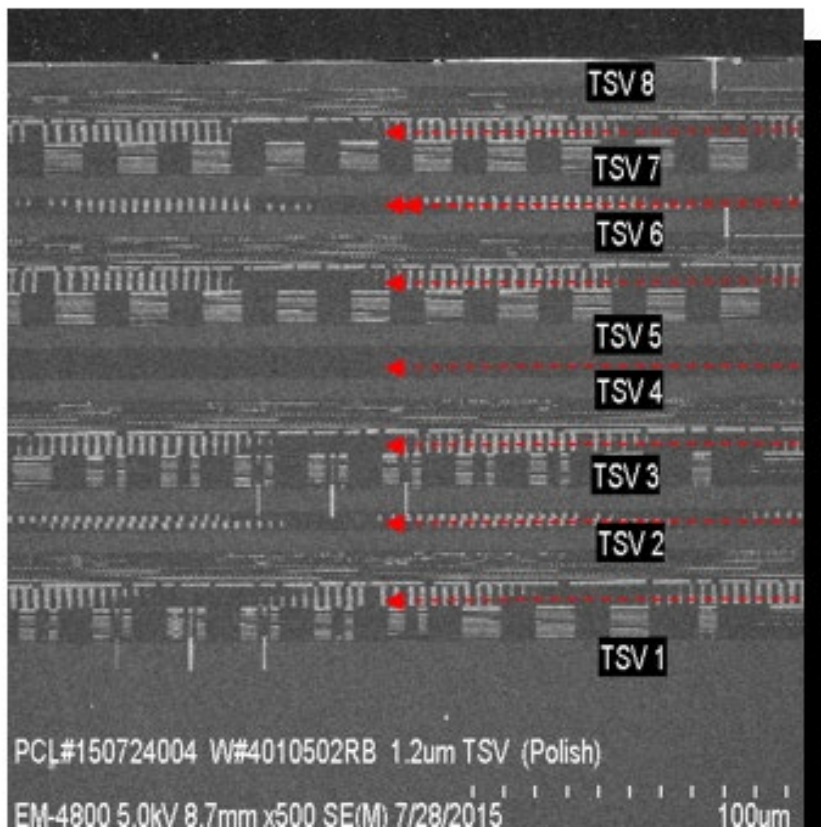

Figure 11: Cross-section of an 8-wafer hybrid bond stack for a 200 mm wafer size (NHanced Image).

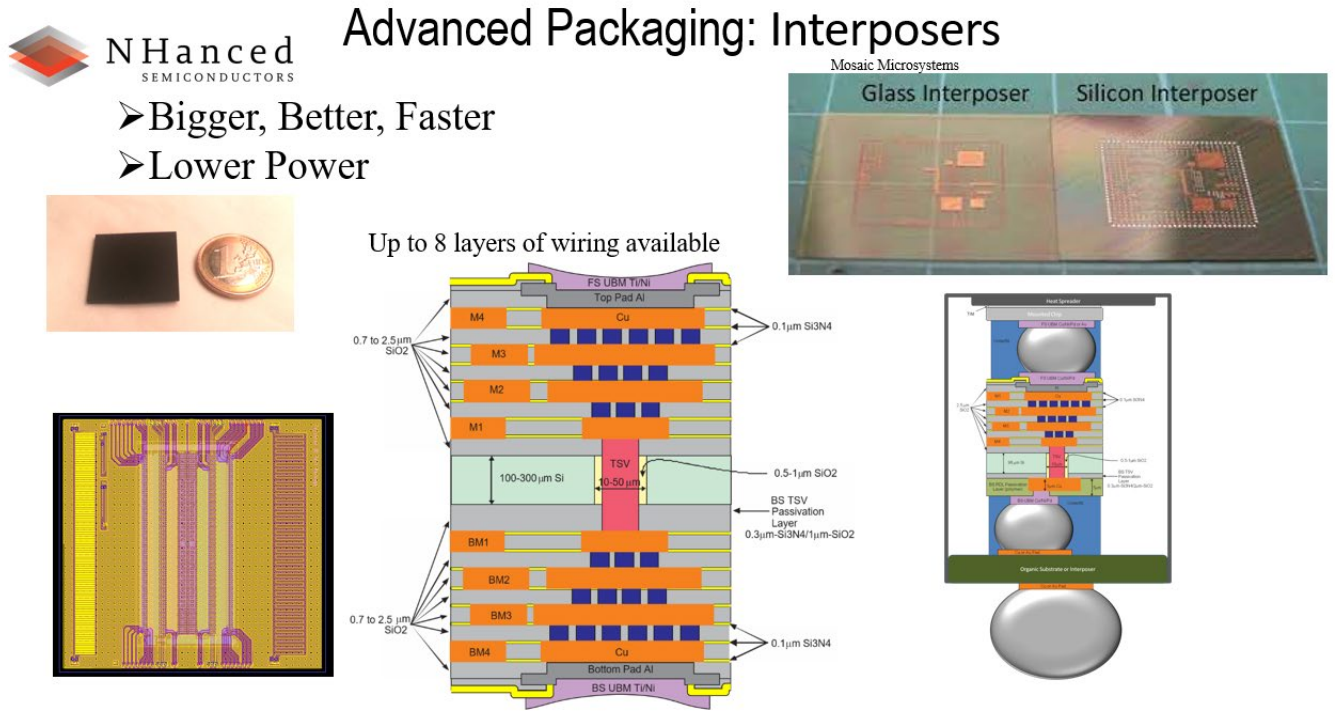

Figure 12: An illustration of a cross-section that has 8 total copper Damascene layers, 4 top and 4 bottom. This capability can be used for both silicon and glass core interposers (NHanced Image).

## V. Metrology

Intrinsic yield loss mechanisms in hybrid bonded chips are primarily driven by the extreme sensitivity of the bonding interface to nanoscale surface and process imperfections. Because hybrid bonding relies on direct dielectric-to-dielectric and Cu-to-Cu contact at very fine pitches, even monolayer-level contamination such as native oxides, moisture, organic residues, or CMP slurry particles can prevent full atomic contact. These imperfections manifest as localized non-bonds, micro-voids, or weak interfaces that directly translate into open circuits or high-resistance connections. In addition, surface topography issues—such as CMP-induced dishing, erosion, and non-uniform dielectric thickness—can prevent uniform contact across the wafer or die, with edge regions often being especially susceptible to

void formation due to non-uniform pressure and polishing effects.

A second major class of intrinsic yield loss mechanisms comes from geometric and mechanical limitations in the bonding process. As interconnect pitch scales down, overlay and alignment errors become increasingly critical, where misregistration on the order of tens of nanometers can result in partial or complete Cu-to-Cu misses. Thermo-mechanical effects such as wafer warpage, die bow, and stress gradients further degrade contact uniformity during bonding, leading to spatially varying bond quality across the interface. Even when initial bonding appears successful, suboptimal Cu surface activation or annealing conditions can limit Cu grain growth and diffusion, resulting in latent reliability issues or early-life resistance drift. Collectively, these mechanisms show that hybrid bonding yield is fundamentally constrained by the interplay of surface chemistry, planarization quality, alignment precision, and mechanical flatness at nanometer-scale tolerances. Table 1 highlights some of the major yield loss mechanisms [4].

Potential Sources of Yield Loss

| # | Defect | Cause |
|---|---|---|
| 1 | Void | Environmental Particles / Manual Handling |
| 2 | Void | Dicing / Edge Quality |
| 3 | Void | Bonder Related: Eject, Flip, Gantry motion |
| 4 | Misalignment | Die Shift / Die Rotation / Deformation |
| 5 | Cracked Die | Thin Die (≤ 50 mm)<br>Unoptimized Singulation or Ejection |
| 6 | Unconnected bond pads | Excessive Recess / Recess Variation<br>Insufficient Anneal |
| 7 | Intrinsic Defects | Defects in metal stack<br>Not related to hybrid bond |

Root Cause Failure Mode Examples

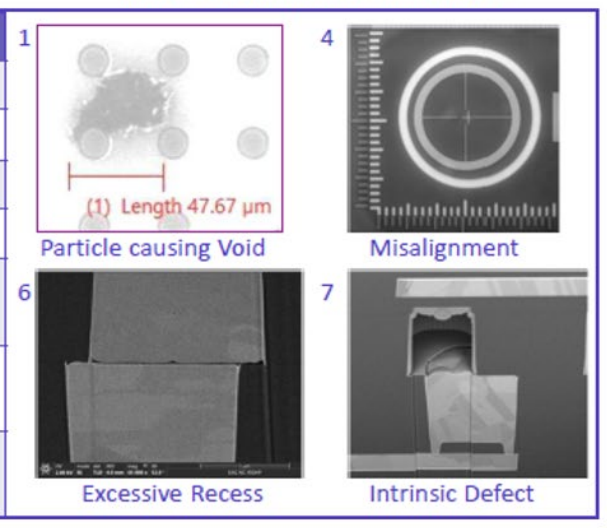

Reference: T. Workman et al., ECTC May 2023

Table 1: Intrinsic yield loss mechanisms for hybrid bonded die.

Reliable high-volume production of ultra-fine pitch hybrid bonds depends on the ability to detect interfacial defects non-destructively, at high throughput, and at the same length scale at which those defects form. The yield-loss mechanisms identified earlier in this paper, including nanoscale contamination, micro-voids and edge voids, copper dishing and dielectric erosion, overlay error, and warpage-induced non-uniform contact, are difficult to capture reliably with any single inspection method. Three-dimensional X-ray microscopy and computed tomography (CT) resolve buried copper structures, through silicon vias (TSVs), and redistribution layers, although both techniques become slow and dose limited as features approach the sub-micron regime. Scanning acoustic microscopy (SAM) is the principal tool for subsurface delamination and void detection at bonded interfaces, yet its images are degraded by noise, limited resolution, and weak contrast at fine pitch [5]. Terahertz (THz) imaging, together with optical and scanning electron microscopy (SEM) methods, adds complementary sensitivity to dielectric integrity and surface defects, although these techniques cannot, on their own, provide the coverage and speed that hybrid-bond yield control requires.

Artificial intelligence (AI) is increasingly used to convert these modalities from manual, operator-dependent measurements into automated and consistent inspection. Deep convolutional neural networks (CNNs) applied to X-ray projection and tomographic data can localize and classify defects automatically, which addresses both the throughput limitation and the variability of human interpretation that constrain X-ray inspection of electronic assemblies. The same principles support automated feature extraction for buried interconnects, such as the detection of vias and conductive traces in reconstructed volumes, a capability that transfers naturally to the inspection of fine-pitch copper pads and TSVs in hybrid-bonded stacks. Because these algorithms operate on the reconstructed three-dimensional volume rather than on a single projection, they are well suited to the stacked, multi-layer geometries that make hybrid bonding difficult to inspect by conventional means.

Acoustic inspection is particularly important for hybrid bonding because its dominant defects, namely interfacial voids, incomplete dielectric fusion, and delamination, are precisely the subsurface flaws that SAM is designed to detect. To overcome the intrinsic noise and resolution limits of SAM, a physics-informed neural network (PINN) framework has been demonstrated that enhances acoustic images by embedding the underlying wave-propagation physics directly into the learning process, improving the visibility of micro-bumps, TSVs, and delamination's while suppressing artifacts [6]. This physics-constrained approach is valuable for trustworthy deployment, because the model is regularized by the governing physics rather than relying on purely data-driven inference, which improves reliability when the technique is qualified for production use. Complementary work uses generative adversarial networks (GANs) to augment limited scanning electron microscopy training data and to improve the recognition of physical features, which helps to mitigate the data-scarcity problem that often constrains the adoption of AI in failure analysis. Taken together, these results point toward a multimodal, AI-driven metrology strategy in which X-ray CT, SAM, THz, and optical or SEM data are acquired non-destructively and analyzed by a common deep-learning engine that detects, classifies, and quantifies the defect populations responsible for hybrid-bond yield loss [7]. Machine-learning methods based on THz imaging extend this multimodal toolkit to dielectric and assurance applications in which X-ray contrast is poor. Surveys of die-attachment and bonding-defect detection further confirm that the combination of optical, radiological, and acoustic sensing with deep learning has become the leading direction for manufacturing-grade defect detection [8]. Figure 13 illustrates the resulting framework, in which the quantified defectivity is returned as a yield model and feed-forward correction to the bonding process, thereby closing the loop between inspection and process control. The central conclusion for ultra-fine pitch hybrid bonding is that metrology must evolve from isolated, manual measurements into an integrated, AI-enabled system that makes inspection fast, repeatable, and increasingly independent of human interpretation [9].

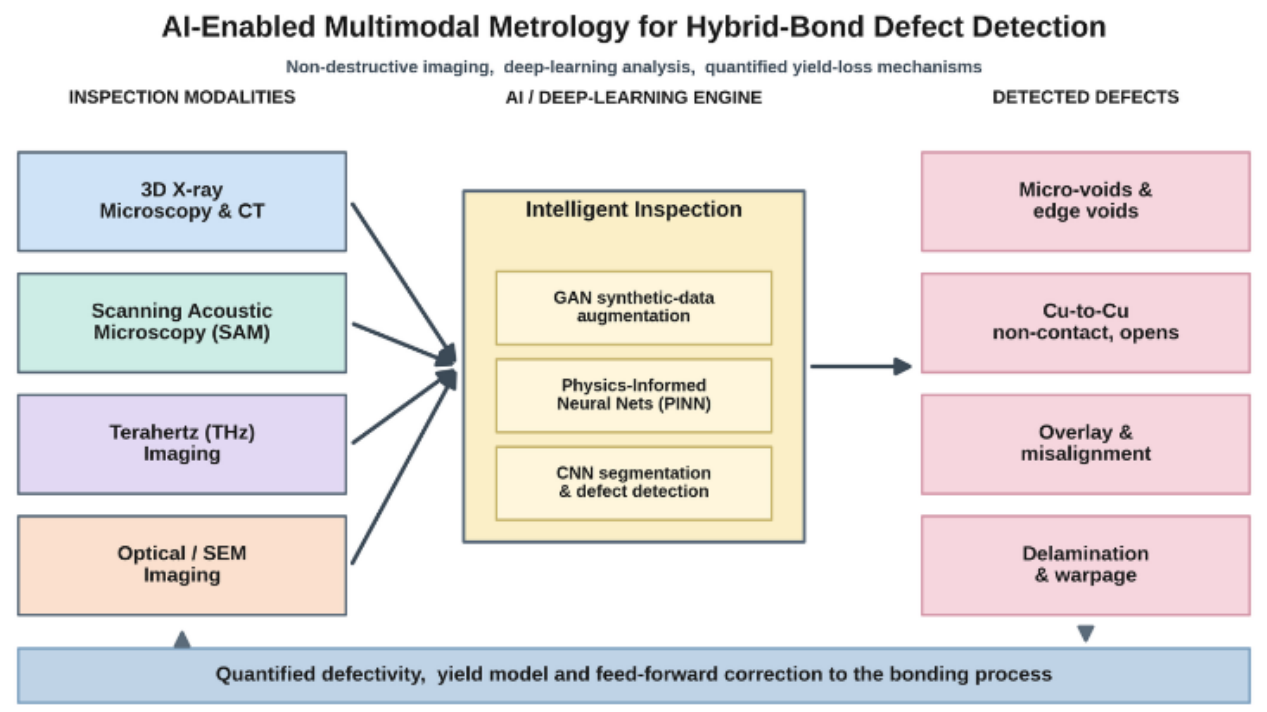

Figure 13: AI-enabled multimodal metrology framework for hybrid-bond defect detection.

Non-destructive imaging modalities feed a common deep-learning engine who's quantified defectivity drives yield feedback to the bonding process. Acronyms used: computed tomography (CT), scanning acoustic microscopy (SAM), terahertz (THz), scanning electron microscopy (SEM), convolutional neural network (CNN), physics-informed neural network (PINN), and generative adversarial network (GAN) [10].

## VI. Discussion

D2W hybrid bonding is increasingly viewed as high-volume manufacturing (HVM)-ready because the process has converged across three tightly coupled domains: (1) sub-100 nm placement control, (2) sub-nanometer surface engineering, and (3) statistically controlled defectivity at scale. On the placement side, modern die bonders now incorporate high-resolution optical alignment systems, multi-point fiducial recognition, and real-time thermal drift compensation that collectively enable overlay approaching the intrinsic requirement window of fine-pitch Cu pads (often ~1–5 µm pitch in advanced implementations). Critical to HVM readiness is that the overlay error distribution is no longer dominated by systematic tool offsets but by bounded stochastic terms (die pick-and-place variation, local chuck non-flatness, and thermal expansion mismatch), which can be modeled and corrected through calibration maps and feed-forward correction. This makes D2W alignment behavior statistically predictable—an essential prerequisite for yield scaling [11].

On the surface and interface side, HVM viability is enabled by CMP processes that now routinely achieve within-wafer non-uniformity control at sub-5 nm scale for dielectric height and sub-1 nm RMS roughness on $SiO_2$. Cu pad formation has also stabilized through damascene integration schemes that tightly control recess/dishing and ensure co-planarity between Cu and dielectric surfaces, which is critical because hybrid bonding relies on simultaneous Cu-Cu contact and dielectric fusion. Plasma activation has become a deterministic step for increasing surface energy and promoting low-temperature covalent bonding of dielectrics while enabling Cu oxide reduction during annealing. Defectivity control has improved via front-end-style contamination management—particle adders are now tracked in the sub-10 nm regime using advanced inspection, and bond void formation is increasingly treated as a yield model parameter rather than an uncontrollable defect mode.

From a system integration perspective, D2W hybrid bonding is also HVM-ready because it aligns naturally with known-good-die (KGD) flows and yield partitioning strategies. Unlike W2W bonding, D2W allows defective die discard prior to bonding, improving effective system yield according to a compounding model:

$$Y_{system} = Y_{die} \cdot Y_{bond} \cdot Y_{wafer}$$

This partitioning is particularly important in advanced nodes where die yield variability dominates cost. Furthermore, thermal-mechanical reliability has been brought under control through improved underfill-less bonding schemes and stress-engineered redistribution layers (RDLs) that mitigate CTE mismatch-induced shear at the Cu-Cu interface. Combined with increasingly standardized process modules (surface prep → plasma activation → precision alignment → thermo-compressive bonding → anneal), D2W hybrid bonding has transitioned from bespoke integration to a repeatable unit process flow, making it compatible with semiconductor HVM environments targeting AI accelerators, 3D memory stacks, and heterogeneous chiplet integration. The readiness of D2W hybrid bonding for HVM rests on placement control, surface engineering, and statistically bound defectivity. What ultimately converts these capabilities into sustained yield is the ability to observe and act on defects in line. AI-enabled inspection provides this link, because it turns non-destructive metrology into a quantitative, closed-loop control system. When deep-learning analysis of X-ray, acoustic, and THz data produces consistent, operator-independent defect counts, bond voids and overlay excursions can be treated as measurable yield-model parameters rather than as uncontrollable defect modes. This statistical predictability is precisely what high-volume manufacturing requires, and it directly addresses the metrology, defect-inspection, and yield-control needs identified at the start of this paper.

A central advantage of the AI approach is multimodal data fusion, since no single modality fully characterizes a hybrid-bonded interface. X-ray CT resolves buried copper geometry, SAM reveals interfacial voids and delamination, and THz and SEM add dielectric and surface sensitivity. By learning across these complementary channels, machine-learning models can classify defects more reliably than any individual technique and at speeds compatible with production cadence, which reduces the manual-inspection bottleneck that has long limited advanced-packaging throughput. Surveys of die-attachment and bonding-defect inspection confirm that combining optical, radiological, and acoustic sensing with deep learning is now the dominant direction for manufacturing-grade defect detection [12].

Several practical considerations must still be managed before these methods can be trusted in qualification and high-volume manufacturing. The first is data scarcity. Because the defect classes of interest are, by design, rare, synthetic-data generation with generative adversarial networks and related augmentation strategies is essential for training robust models without waiting for large defective populations to accumulate. The second is trustworthiness and explainability. For an inspection decision to gate a costly stacked device, the output of the model must be interpretable and physically grounded, and physics-informed networks that embed imaging or wave-propagation physics in the architecture offer a promising route to models that are both accurate and defensible during process qualification. The third is integration with the broader supply chain and assurance flow, in which the same imaging and AI infrastructure that detects manufacturing defects also support authenticity and reliability verification across heterogeneous, multi-sourced components [13].

The natural end state is a closed-loop, AI-enabled manufacturing system, illustrated in Figure 14, in which multimodal inspection feeds machine-learning classification, which in turn drives statistical process control and a digital-twin representation of the bonding process, and finally provides feed-forward correction to the bonder. Organized as layers of data, model, decision, and control, the same capability provides a roadmap for increasing autonomy and yield control as the technology scales [14]. Within this framework, AI-enabled X-ray, acoustic, and THz inspection is not a downstream quality check but an integral, real-time element of the process. It is this capability that makes ultra-fine pitch D2W hybrid bonding statistically predictable, less dependent on human interpretation, and therefore genuinely ready for high-volume manufacturing of AI accelerators, three-dimensional memory, and heterogeneous chiplet integration.

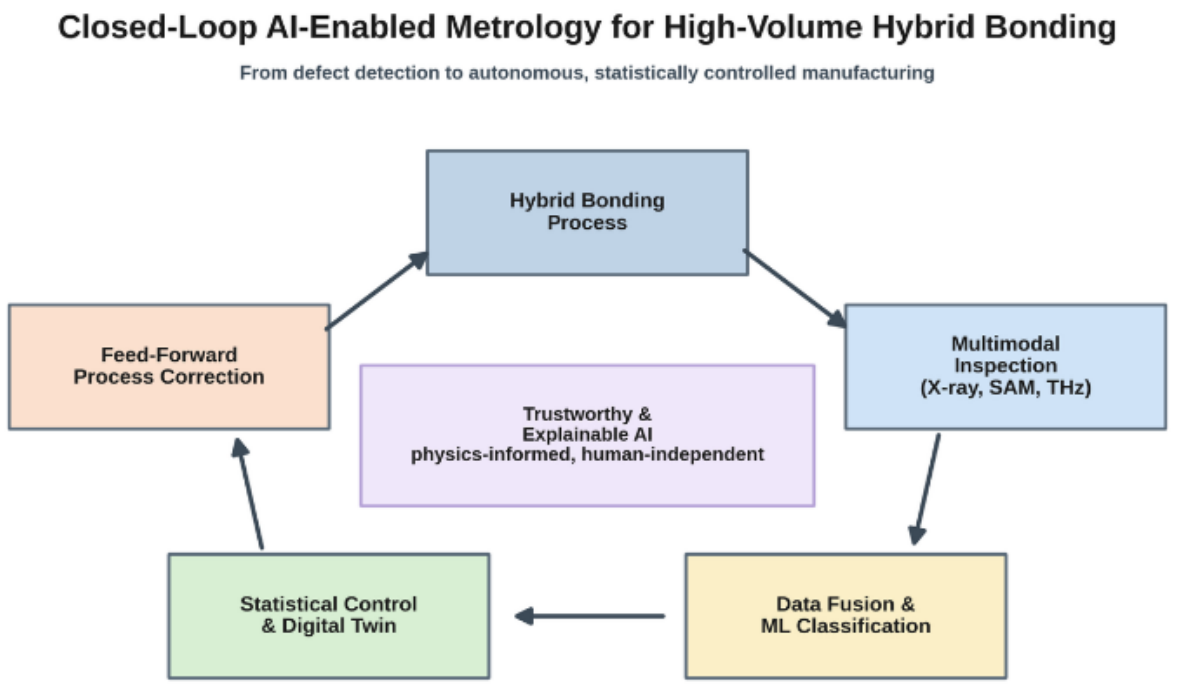

Figure 14: Closed-loop AI-enabled metrology for high-volume hybrid bonding. Inspection, machine-learning classification, statistical control with a digital twin, and feed-forward correction form a continuous loop governed by trustworthy, explainable AI.